\documentclass[letterpaper, 10 pt, conference]{ieeeconf}  

\IEEEoverridecommandlockouts                              

\usepackage{graphicx} 
\usepackage{xcolor}
\usepackage{comment}
\usepackage{color}
\usepackage[flushleft]{threeparttable}
\usepackage{makecell}

\usepackage{fancyhdr}
\fancypagestyle{firstpage}{%
    \chead{This paper has been accepted for publication at the 2024 International Symposium on Medical Robotics.}
    }

\title{\LARGE \bf
Towards Biomechanical Evaluation of a Transformative Additively Manufactured Flexible Pedicle Screw for Robotic Spinal Fixation 
}

\author{Yash Kulkarni$^{1}$, Susheela Sharma$^{1}$, Jordan P. Amadio$^{2}$, and Farshid Alambeigi$^{1}$
\thanks{*This work is supported by the National Institute Of Biomedical Imaging and Bioengineering of the National Institutes of Health under Award Number R21EB030796.}
\thanks{$^{1}$Y.~Kulkarni, S.~Sharma, and F.~Alambeigi are with the Walker Department of Mechanical Engineering and the Texas Robotics  at the University of Texas at Austin, Austin, TX, 78712, USA. Email: \{kulkarni.yash08, sheela.sharma\}@utexas.edu. \{farshid.alambeigi\}@austin.utexas.edu}%
\thanks{$^{2}$J.~P.~ Amadio is with the Department of Neurosurgery, The University of Texas Dell Medical School, TX, 78712. }
}

\begin{document}

\maketitle
\thispagestyle{firstpage}
\pagestyle{empty}

\begin{abstract}
Vital for spinal fracture treatment, pedicle screw fixation is the gold standard for spinal fixation procedures. Nevertheless, due to the screw pullout and loosening issues, this surgery often fails to be effective for  patients suffering from osteoporosis (i.e., having low bone mineral density). These failures can be attributed to the rigidity of existing drilling instruments and  pedicle screws forcing clinicians to place these implants into the osteoporotic regions of the vertebral body. To address this critical issue, we have developed a steerable drilling robotic system and evaluated its performance in drilling various J- and U-shape trajectories. Complementary to this robotic system, in  this paper, we propose design, additive manufacturing, and biomechanical evaluation of a transformative flexible pedicle screw (FPS) that can be placed in pre-drilled straight and curved trajectories. To evaluate the performance of the proposed flexible implant, we designed and fabricated two different types of FPSs using the direct metal laser sintering (DMLS) process. Utilizing our unique experimental setup and   ASTM standards, we then performed various pullout experiments on these FPSs to evaluate and analyze their biomechanical performance implanted in straight trajectories. 
\end{abstract}

\section{INTRODUCTION}
Osteoporosis is a critical health concern responsible for 2 million broken bones annually and 57 billion dollars in operating cost in The United States alone \cite{Lewiecki2019HealthcarePC}. Among osteoporotic fractures, vertebral compression fractures are the most prevalent with more than 1.4 million global occurrences \cite{Johnell2006AnEO}. In cases where nonsurgical methods prove to be insufficient to fix vertebral compression fractures, minimally invasive surgical interventions such as spinal fixation (SF) may be required. In SF, two (or more) vertebra are fused together using a biocompatible implant. This surgery is currently performed using a rigid drilling instrument to create linear trajectories through the rigid cortical bone corridors of the vertebra pedicles. Subsequently, rigid pedicle screws (RPS) are inserted through the corridors, fixating within the porous cancellous bone of the vertebral body \cite{Rometsch2020ScrewRelatedCA}. Finally, the two RPS are connected together using locking rods, restoring stability to the spine and eliminating painful motion. 

Despite RPS fixation being considered the gold-standard for SF surgery, significant shortcomings persist in the current procedure. The most important being the loosening and pullout of the RPS with a reported incidence as high as 22-50\% even in cases with healthy bone mineral density (BMD) (e.g., \cite{Rometsch2020ScrewRelatedCA,Weiser2017InsufficientSO}). Furthermore, the RPS fixation's stability is barely sufficient for an osteoporotic vertebral body  (i.e., BMD $<$ 80 mg/cm$^3$) \cite{Inceoglu,Polly1998RevisionPS}, leading to fixation failure and revision surgery \cite{Wittenberg1991ImportanceOB,Weiser2017InsufficientSO}. 

	 \begin{figure}[t!]
		\centering 
		\includegraphics[width=1\linewidth]{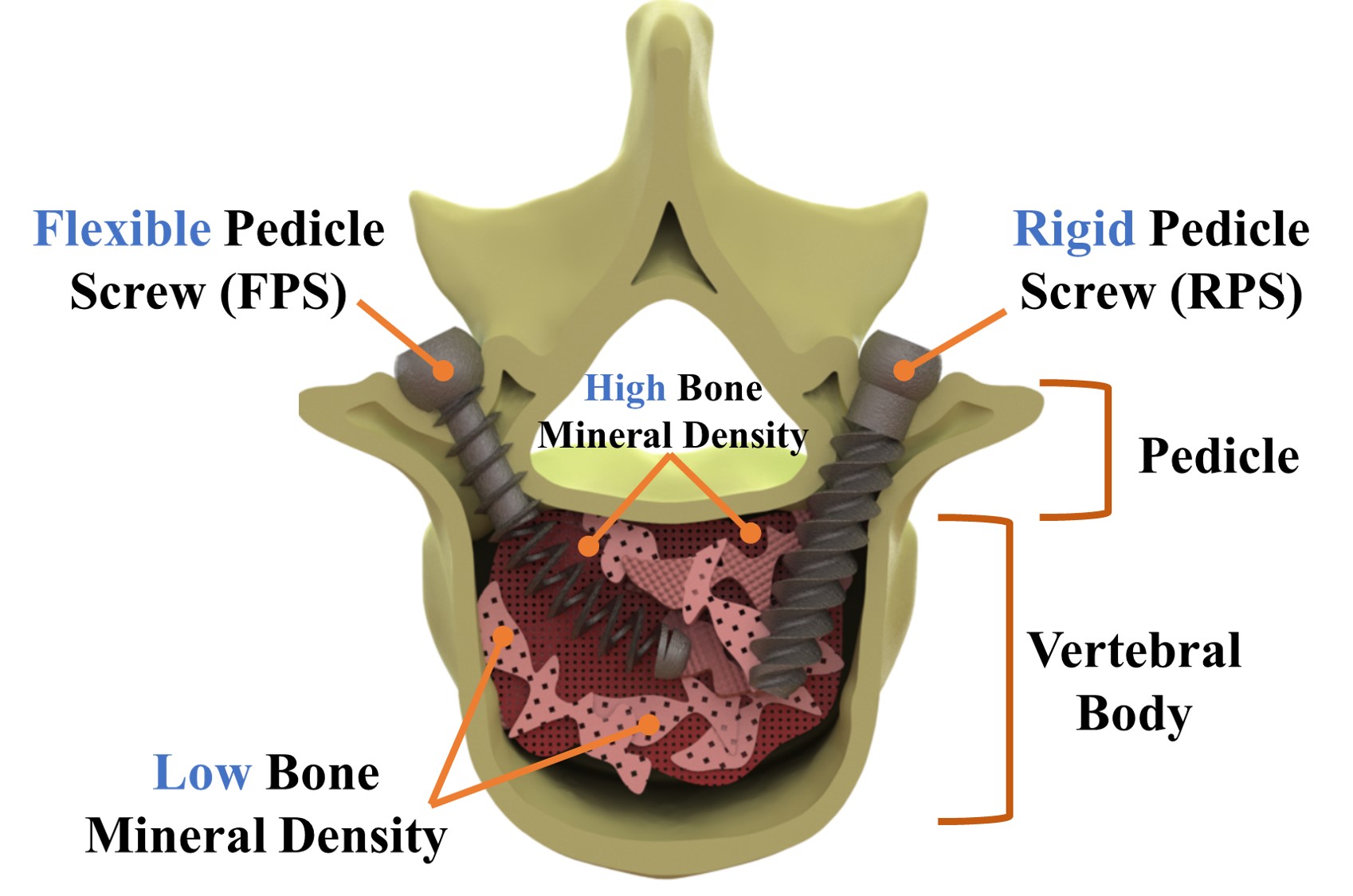}
		\caption{Conceptual illustration of the proposed FPS. Left: The proposed FPS is fixed into a curved trajectory within a spinal vertebra. Notably, it is shown to be avoiding areas of low BMD. Right: A RPS fixed into areas of high BMD.}
            \vspace{-5mm}
		\label{fig:Rigid_Flex_Concept}
	\end{figure}

Numerous approaches have been explored to address these complications \cite{Rometsch2020ScrewRelatedCA}, including (1) improving the design of the RPS through changing its parameters (e.g., thread diameter \cite{Talu2000PedicleSS} and pitch \cite{Mehta2012BiomechanicalAO}), (2) creating novel screws capable of anchoring into bone by opening outward \cite{Shea2014DesignsAT}, and (3) exploring robotic systems and navigation techniques to improve RPS fixation \cite{Sefati2021ASR}. However, all of these methods still prove to be ineffective in solving the screw loosening problem and continue to lead towards revision surgeries \cite{Phan2015CorticalBT}. A review of the literature \cite{Alambeigi2018InroadsTR,alambeigi2019use,Sharma2023TowardsBD,Sharma2023ACT,Alambeigi_2017,bakhtiarinejad2020biomechanical} reveals that the main cause of these shortcomings is due to current rigid instruments constraining the location of spinal fixation implant placement. As illustrated in Fig. \ref{fig:Rigid_Flex_Concept}, the current RPS design can force the screw fixation into areas of low BMD and lead to an increased risk of screw loosening \cite{Rometsch2020ScrewRelatedCA}. 

 Typically pedicle screws tend to loosen in the vertebra overtime due to daily activity; however, a review of the literature (e.g., \cite{Wu2023PulloutSO,Abshire2001CharacteristicsOP}) demonstrates the most common way to evaluate the fixation strength of a pedicle screw is through a pullout strength test. While the method of failure produced by the pullout strength test is uncommon, the test itself provides a common way to compare pedicle screws \cite{Abshire2001CharacteristicsOP}.

\begin{figure}[t!]
    \centering
    \includegraphics[width=0.85\linewidth]{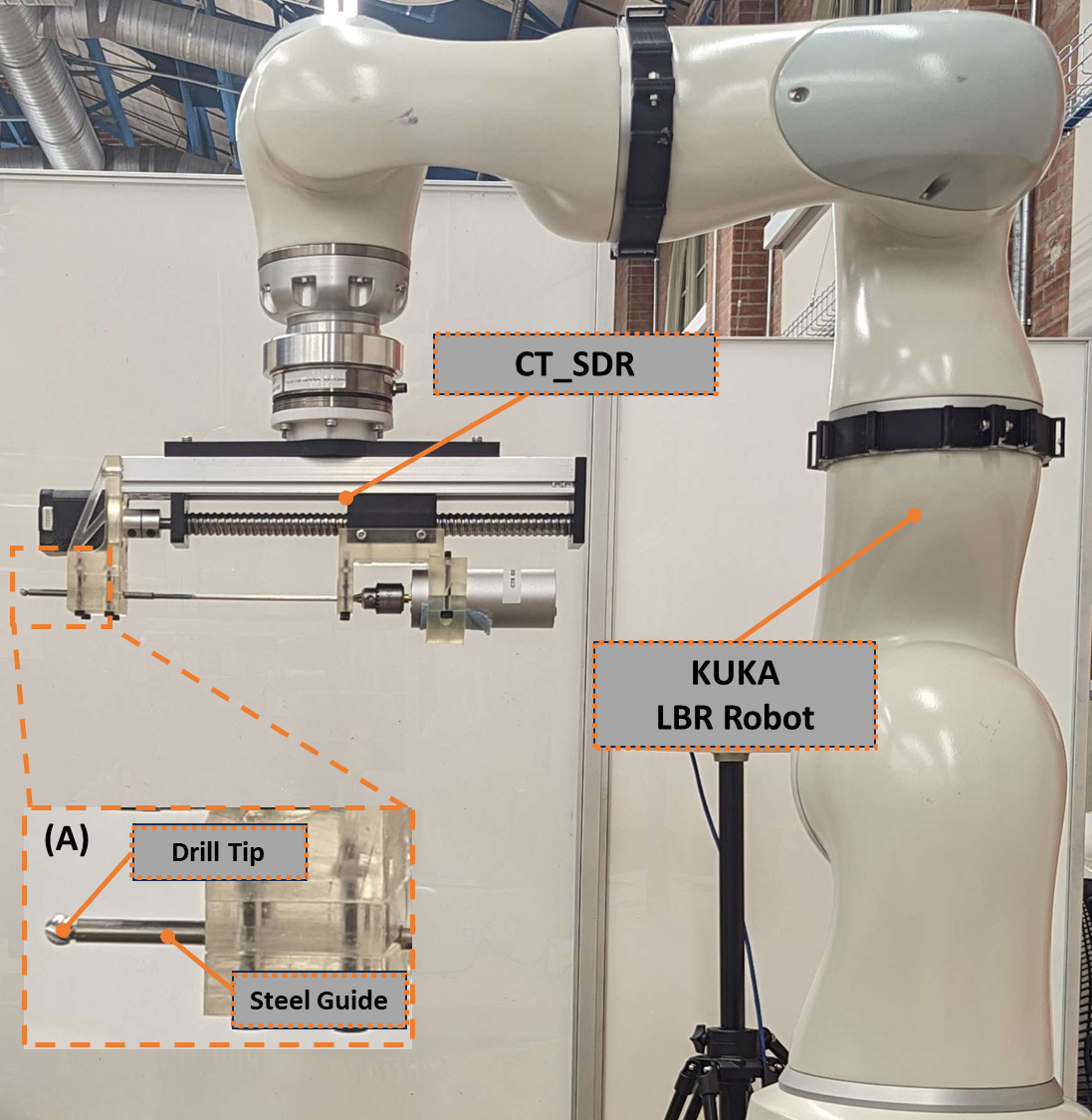}
    \caption{Experimental Set-Up of the CT-SDR System including the CT-SDR and KUKA LBR. (A): A detailed view of the flexible cutting tool including the steel guide and the drill tip.}
    \vspace{-5mm}
    \label{fig:Robot}
\end{figure}

This paper addresses the pullout and loosening limitations of existing RPSs by proposing the design, manufacturing, and evaluation of a novel flexible pedicle screw (FPS). As conceptually shown in Fig. \ref{fig:Rigid_Flex_Concept}, we will introduce the design for a FPS capable of safely exploiting the natural geometry of a patient's vertebrae. The proposed FPS can morph within and tap a pre-drilled curved tunnel created by a novel concentric tube steerable drilling robot (CT-SDR) \cite{Sharma2023ACT,Sharma2023TowardsBD} as shown in Fig. \ref{fig:Robot}. Furthermore, to evaluate the proposed FPS design, we use a direct metal laser sintering (DMLS) process to fabricate two different FPS in titanium.  Despite the capability of FPS in morphing into curved trajectories, in this work, we solely  validated the fixation strength of the FPS implanted into a straight pre-drilled  trajectory using our robotic system. To perform the test and evaluate the performance of FPS,  we utilized ASTM standard \cite{ASTMF543} defining pullout strength test and guidelines. 

\section{FPS for Internal Fixation}
As conceptually shown in Fig. \ref{fig:Rigid_Flex_Concept}, an RPS  consists of a rigid body  constraining it to  follow a limited linear or straight trajectory leading to the aforementioned problems in the previous Section. Therefore, the FPS we propose must deviate from this design in order to freely bend in a pre-drilled J-shape trajectory already drilled using a steerable drilling robot (e.g., \cite{Alambeigi2018InroadsTR,alambeigi2019use,Sharma2023TowardsBD,Sharma2023ACT,Sharma2024ABR,Sharma_ismr2024}) and fixate in areas of high BMD regions within the vertebral body. 
Building on the previous work by Alambeigi et al \cite{Alambeigi2018InroadsTR,alambeigi2019use}, we propose an FPS design with the following characteristics to address the aforementioned needs:

\subsection{Semi-Flexible Semi-Rigid Body}
Ensuring structural integrity of the vertebra while promoting bone growth is critical for overall patient safety. In order to ensure the stability of the vertebra, the semi-flexible semi-rigid combination plays a critical role. The semi-flexible (as marked by \raisebox{.5pt}{\textcircled{3}} on Fig. \ref{fig:Screw_Design}), semi-rigid (as marked by \raisebox{.5pt}{\textcircled{4}} on Fig. \ref{fig:Screw_Design}) combination ensures the screw can enhance support in the pedicle region of the screw while still being capable of bending to reach high BMD areas. Furthermore, the gaps between the threads in the flexible region allow for bone growth through the screw providing an opportunity for enhanced bio-integration with the vertebra.

\subsection{Rounded Head}
Rigid pedicle screw breaching is a common concern. To ensure that no breaching occurs between pedicle screw tip and the curved trajectory created by the CT-SDR, a rounded head tip is designed (as marked by \raisebox{.5pt}{\textcircled{1}} on Fig. \ref{fig:Screw_Design}). This rounded tip forces the screw body to deflect when it intersects the curved trajectory walls, guaranteeing a safe interaction for the patient.

\subsection{Self-tapping Threads}
\begin{figure}[t!] 
		\centering 
		\includegraphics[width=1\linewidth]{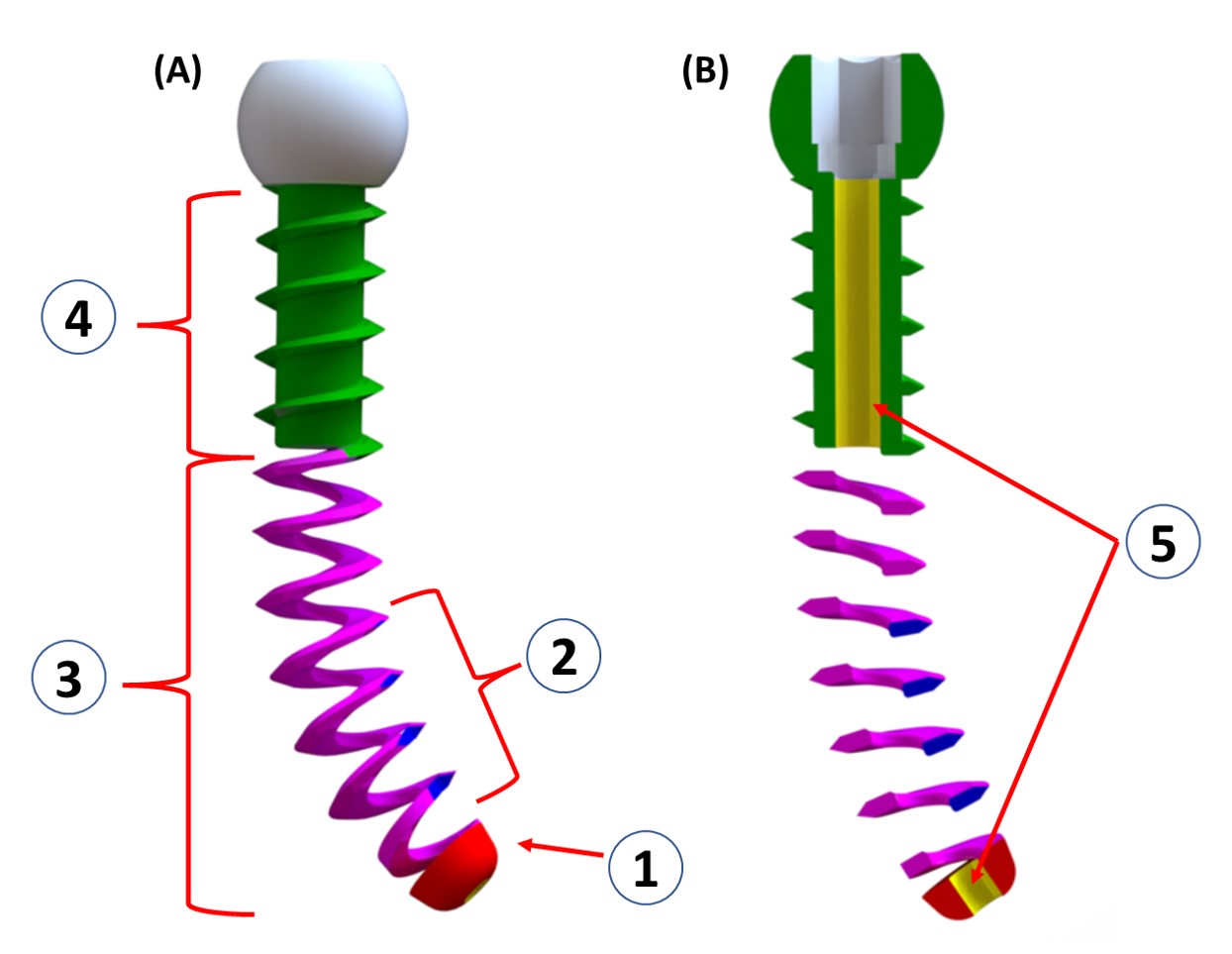}
		\caption{Theoretical design of the FPS emphasizing critical design features based on patient and surgeon necessities.\textcircled{\raisebox{-0.9pt}{1}} represents the rounded head of the screw. \textcircled{\raisebox{-0.9pt}{2}} represents the self-tapping threads of the screw, with \textcircled{\raisebox{-0.9pt}{3}} and \textcircled{\raisebox{-0.9pt}{4}} representing the flexible and rigid parts of the screw, respectively.  \textcircled{\raisebox{-0.9pt}{5}} represents the cannulated region of the screw. (A) illustrates the side view of the full FPS screw with (B) representing the cross-section of the FPS.}
            \vspace{-5mm}
		\label{fig:Screw_Design}
	\end{figure}

Self-tapping threads (as marked by \raisebox{.5pt}{\textcircled{2}} on Fig. \ref{fig:Screw_Design}) are also a vital component of the flexible region of the screw. The self-tapping threads play a critical role in ensuring the FPS enters the curved trajectory as smoothly as possible while reducing the overall operating cost and time for both the patient and surgeon. 

	 \begin{figure}[t!] 
		\centering 
		\includegraphics[width=1\linewidth]{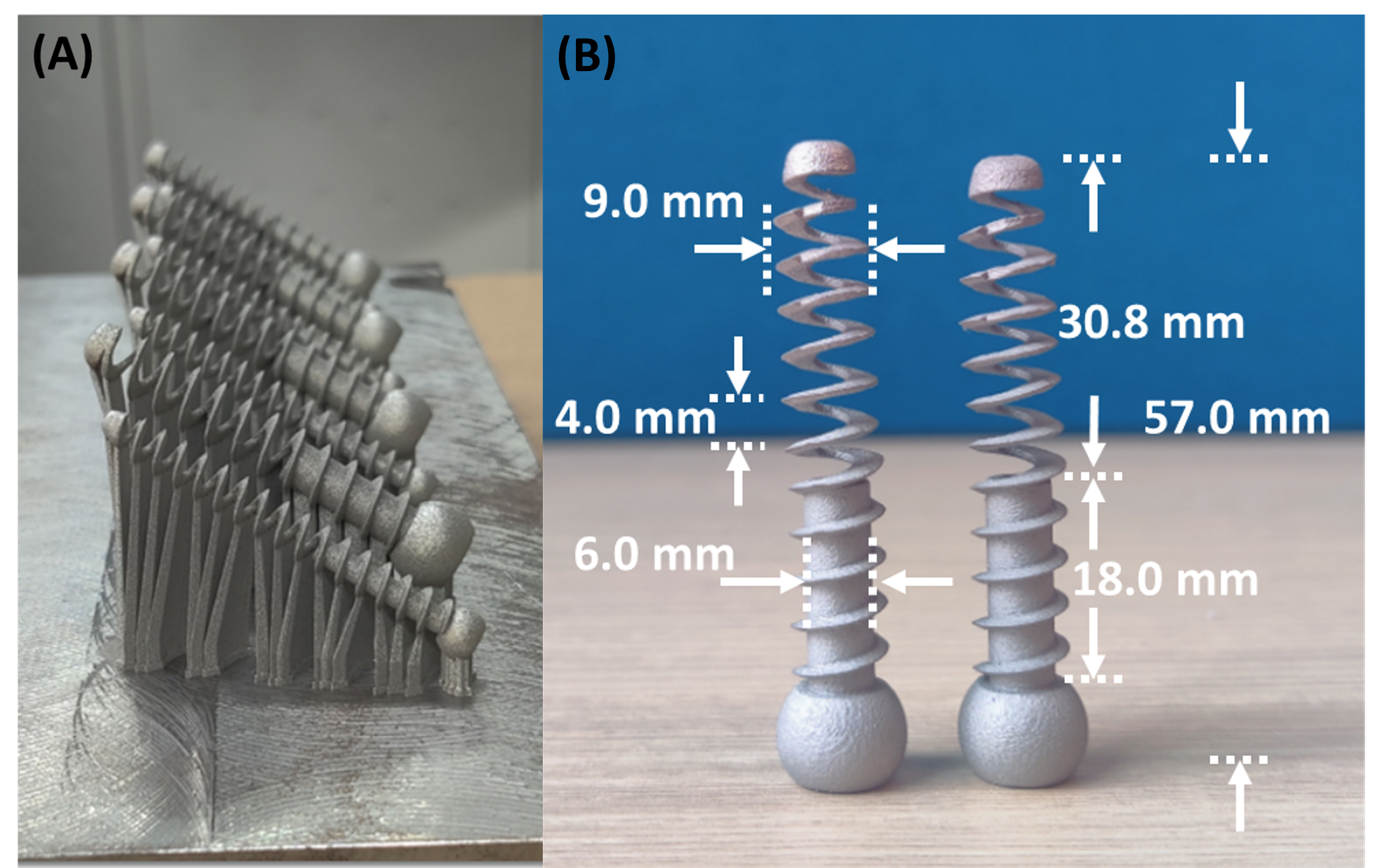}
		\caption{Metal 3D printed flexible pedicle screw. (A) Two different sized flexible pedicle screws fabricated on the build plate before post-processing.  (B) The 9 mm OD-6 mm ID screw after post-processing had been done including sandblasting. Critical dimensions for the screw are annotated on the image.}
            \vspace{-5mm}
		\label{fig:AMBuild}
	\end{figure}

\subsection{Cannulated Region}
The cannulated region (as marked by \raisebox{.5pt}{\textcircled{5}} on Fig. \ref{fig:Screw_Design}) plays a critical role in ensuring patient safety by providing a method to enhance the FPS strength and vertebra integrity after the insertion process is complete. The cannulated region provides an opening for the potential injection of bone cement (e.g., polymethylmethacrylate (PMMA)), if necessary. This injection of bone cement further enhances the FPS fixation inside the vertebra.

\section{Additive Manufacturing of the FPS}
Based on conceptual design framework, geometry of an L3 vertebra \cite{Zindrick1986ABS},  the dimensions of the fabricated CT-SDR \cite{Sharma2023ACT,Sharma2023TowardsBD} (shown in Fig. \ref{fig:Robot}), we designed two different FPSs with a 9.0 mm outer diameter (OD), 6.0 mm inner diameter (ID), and a pitch of 3.0 mm and 4.0 mm. As shown in Fig. \ref{fig:AMBuild}, the overall length of the FPS with 4.0 mm pitch  is 57.0 mm with an 18.0 mm rigid region and a 30.8 mm flexible region and 12 threads. The overall length of the FPS with 3.0 mm pitch  is 56.7 mm with an 18.0 mm rigid region and a 30.3 mm flexible region and 15 threads. Of note, without loss of generality, the OD and ID where guided by the CT-SDR ability to drill an average hole size 8 mm in diameter. The fabricated FPS allows us to evaluate and verify the design based on current tools. The first 4 threads of the FPS are self-tapping while the rest are sharp V-threads. It is worth emphasizing that these dimensions can readily be modified depending on the utilized drilling system and geometry of vertebra.

To ensure biocompatibility of FPS and similar to the current FDA-approved  implants, the FPS is made from commercially sourced Titanium Ti-6Al-4V granulates from EOS (Krailling, Germany). The build was prepared using Renishaw's complementary QuantAM software (Leuven, Belgium). The Renishaw AM 250 Laser Melting System (Leuven, Belgium) was used to additive manufacture the Ti-6Al-4V granulates together with a layer thickness of 30 micrometers on a complementary titanium build plate. Figure \ref{fig:AMBuild}-A illustrates the FPS on the titanium build plate right after the additive manufacturing process. 
Figure \ref{fig:AMBuild}-B shows the 4 mm pitch screw after post-processing is ne.

\section{Experimental Setup and Procedure}
As the FPS is expected to strongly fixate within the vertebra body, validating its pullout strength is crucial to verify the overall efficacy of the novel medical device. The pullout test is an standard test to evaluate the performance of PSs and  measure the overall axial tensile force needed to remove the FPS from a simulated bone sample (Sawbone, Pacific Research Company) it is implanted in \cite{etin2021ExperimentalIO, ASTMF543}. 

	 \begin{figure}[t!] 
		\centering 
		\includegraphics[width=1\linewidth]{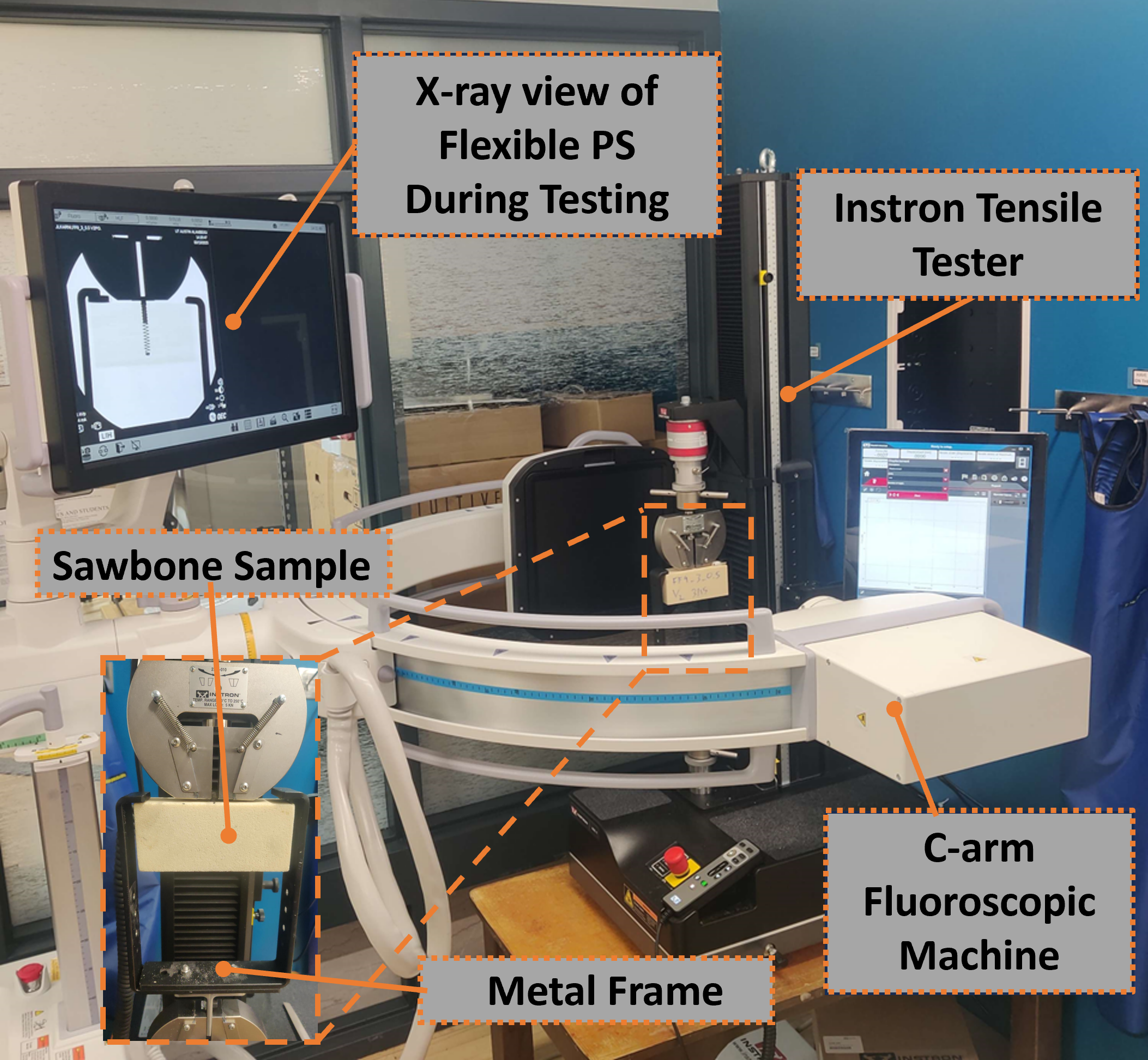}
		\caption{Experimental set-up for pedicle screw pullout testing. The experimental set-up includes a X-Ray C-Arm machine, an Instron Tensile Testing Machine, and PCF-10 Sawbone samples fitted within a metal frame. It also has a zoomed in view of the Sawbone sample interacting with the Instron machine during the pullout process.}
		\label{fig:Setup}
            \vspace{-5mm}
	\end{figure}

 	 \begin{figure*}[t] 
		\centering 
\includegraphics[width=1\linewidth]{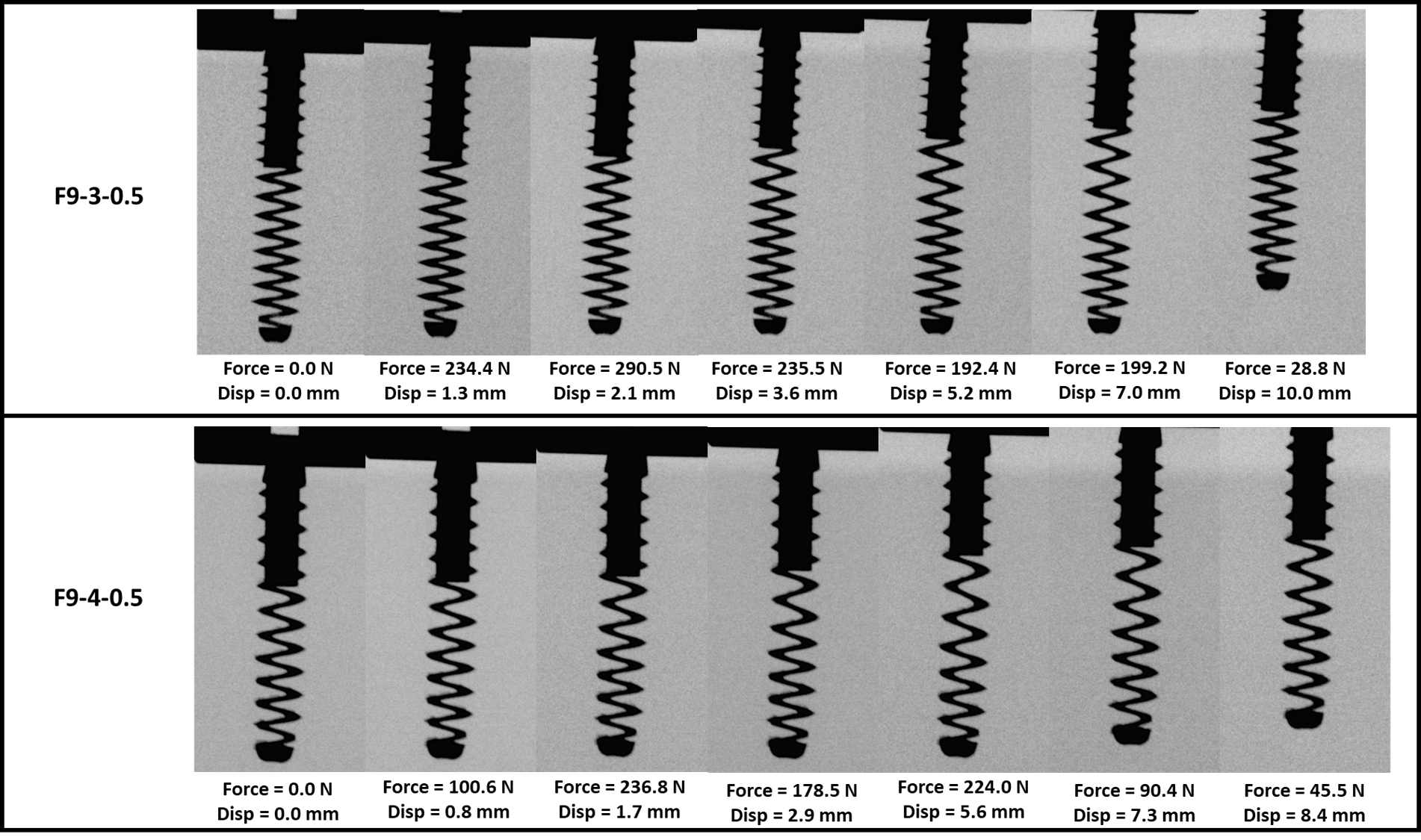}
		\caption{X-Ray images illustrating the 9 mm OD - 6 mm ID FPS as they are pulled out of their Sawbone samples with corresponding force and displacement listed below. Top: The 9 mm OD 3 mm pitch FPS as it gets pulled out at 0.5 mm/min. Bottom: The 9 mm OD 4 mm pitch FPS as it gets pulled out at 0.5 mm/min.}
            \vspace{-5mm}
		\label{fig:Xray}
	\end{figure*}

The set-up for performing the pullout test experiment and capturing the load-displacement  relationship is illustrated in Fig. \ref{fig:Setup}. Two shelf brackets were initially designed and conjoined with two L-shaped brackets to create a rigid metal frame. The rigid metal frame is then mounted into the Instron Tensile Tester (Instron, Massachusetts, USA). The Instron Tensile Tester has a resolution of 0.0001 N. PCF 10 Sawbone phantoms (Sawbones; Pacific Research Laboratories, Washington, USA) are used conforming to ASTM F1839 \cite{ASTMF1839} to ensure conformity of mechanical properties between samples. Of note, as shown by \cite{etin2021ExperimentalIO}, Sawbone samples with PCF 10 density replicate  an osteoporotic bone that result in the most spinal fixation failure cases ($>$90\% \cite{Weiser2017InsufficientSO}) due to pullout of utilized RPSs. Therefore, in this study, we only performed our tests on these samples, to evaluate our proposed flexible implants for such situations.  The Sawbone samples are cut to an appropriate size to match the metal frame using a bandsaw, and straight insertion holes are created using a standing drill press with a rigid drill bit.  Considering that a pilot hole up to 85\% of the OD of the pedicle screw has no affect on pullout strength, \cite{Heidemann1998InfluenceOD,Einafshar2022EvaluationOT}, a 6 mm straight drill bit was used to ensure this recommendation is met. A 6 mm drill bit tended to create a 6.5 mm pilot hole on average. The FPS is then fully inserted into the Sawbone phantom sample without any pre-tapping. Following guidelines of ASTM F543-23 \cite{ASTMF543}, the FPS head is then placed into the head of the Instron Tensile Tester. Of note, as mentioned, in this paper, we solely focused on evaluating the performance of the designed and fabricated FPSs when implanted into straight drilled trajectories. Therefore, without loss of generality, instead of using our robotic system, we used a conventional rigid drill bit to create a straight pilot hole. 
 
Once the FPS is securely clamped into the Instron machine, a displacement load of 0.5 mm/min is applied until the overall pullout strength falls below 10 Newtons. The experiment was repeated twice for both screws and then averaged. This was done to ensure the structural integrity of the FPS.

As shown in Fig. \ref{fig:Setup}, we also synched a C-Arm Xray machine (OEC One CFD, GE Healthcare) with the tensile tester to simultaneously  take real time images of the pullout process and see through the sawbone samples and analyze deformation of FPS during the process. Figure \ref{fig:Xray} are the outputs of the X-ray images taken through the pullout tests.
These real time images were then inputted into Matlab-based Ncorr software \cite{Blaber2015NcorrO2,Harilal2014AdaptationOO} to measure the mechanical deformations of individual threads of the FPS.

For the ease of presentation, from now on in the paper,  we use the following conventions to refer to the fabricated FPSs: FA-B-C. In this representation, F represents an FPS and A, B, and C numbers indicate  the OD, pitch, and  pullout speed during the tensile tests, respectively.

\section{Results and Discussion}
It is critical design requirement that the FPS can provide strong internal fixation within the vertebra to prevent screw loosening and pullout. Holistically, the mechanical behavior of the FPS can be investigated using traditional pullout techniques. 
On a individual thread level, deformation analysis plays a critical role in understanding the  thread deformation behavior. Overall, the performed pullout experiments provided us with an in-depth holistic understanding of the fixation strength of the FPS  while also providing us insight into the deformation behavior of the  flexible region. 

 	 \begin{figure*}[t] 
		\centering 
		\includegraphics[width=0.8\linewidth]{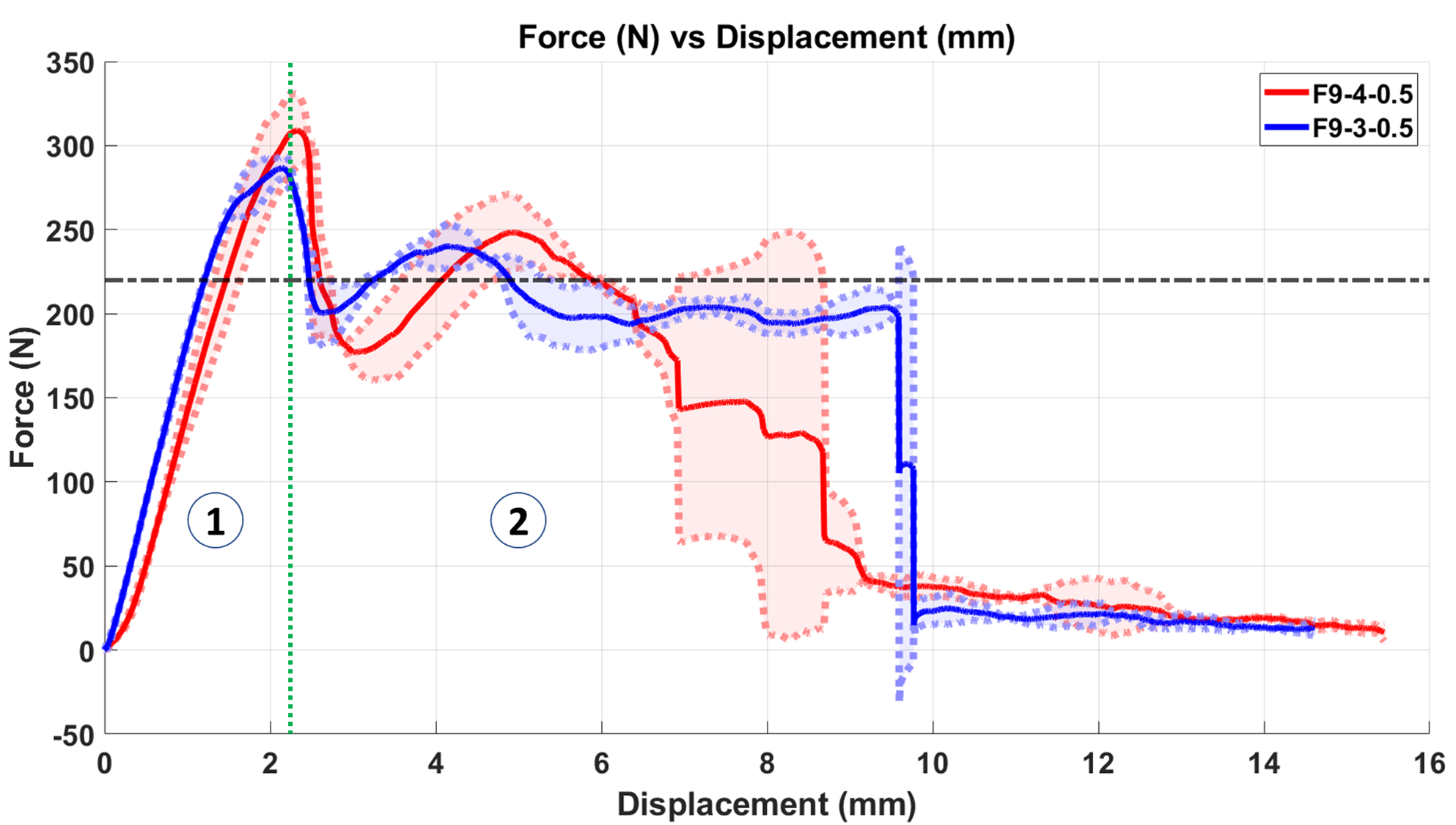}
		\caption{Averaged FPS pullout measurements for the two different FPS (3 mm pitch and 4 mm pitch) with force against displacement. The red and blue indicates the average values of the experiments with the black indicating the minimum acceptable standard for pullout test. The green value indicates the divide between the typical linear rising region common with pedicle screws (marked with \raisebox{.5pt}{\textcircled{1}}) and the prolonged fixation zone (marked with \raisebox{.5pt}{\textcircled{2}}).  The red and blue shade indicating the standard deviations of the experiments.}
            \vspace{-5mm}
		\label{fig:Load-Disp}
	\end{figure*}

\subsection{Pullout Strength Test}

Figure \ref{fig:Load-Disp} illustrates the results for the performed pullout tests. Using the MATLAB \emph{ShadedErrorbar} function, the standard deviation between the trials are added as shaded regions to the figure to show variability between the tests. Furthermore, there are four major lines on the graph marked in red, blue, black, and green. The red line indicates the average force-displacement curve for the 4 mm pitch FPS, the blue line is the average force-displacement curve for the 3 mm pitch FPS, and the black line indicates the ASTM F1839 \cite{ASTMF1839} required minimum force threshold for a pedicle screw in PCF 10 Sawbone. The green line is used to indicate the boundary between the 2 regions seen on Fig. \ref{fig:Load-Disp}. While the overall load-displacement curve is the same between the two FPSs, it marks a stark contrasts with the load-displacement curve of a typical RPS. 

The first region (as marked by \raisebox{.5pt}{\textcircled{1}} on Fig. \ref{fig:Load-Disp}) of the FPS is similar to that of a RPS with an linear rise to a maximum peak \cite{etin2021ExperimentalIO}. After this maximum peak is reached; however, the RPS tends to quickly drop till it reaches 0 N \cite{etin2021ExperimentalIO}.  With the FPS, we see a slight drop occurring after the maximum peak is reached before the threads in the flexible region of the FPS activate. As shown in Fig. \ref{fig:Xray}, this activation leads to a spring-like effect to occur due to the geometry of the flexible region of the FPS and the elastic energy stored in the flexible threads. This activation leads to an prolonged fixation to occur (as marked by \raisebox{.5pt}{\textcircled{2}} on Fig. \ref{fig:Load-Disp}) and prevents the immediate loss of fixation strength typically seen with a RPS. This prolonged fixation zone (PFZ) is a critical improvement to the current standard as it allows for the FPS to be fixated inside the bone longer and preventing the immediate slippage of the screw after the maximum peak is reached as typically seen with RPSs.

Furthermore, the pitch also influences the peak pullout strength as well as maximum displacement till failure as shown in Fig. \ref{fig:Load-Disp}. The greater the pitch, the stiffer the FPS tends to be leading to a higher peak pullout strength but also leading to quicker failure of the FPS. As expected, the F9-4-0.5 reached a higher initial peak force of 309.00 N due to being stiffer while the F9-3-0.5 had a initial maximum peak of 286.75 N.
In exchange, the F9-4-0.5 experienced its first major force drop-off in the PFZ around 6.9 mm while the F9-3-0.5 experienced its drop-off around 9.6 mm.  Overall, both FPSs exceed the minimum requirements set by ASTM (i.e., 220 N \cite{ASTMF1839}) to be considered as an acceptable pedicle screw by a wide margin. These features indicate  the potentials of FPS in providing a better fixation compared with existing RPSs and the current fixation methods.

\subsection{Thread Deformation Analysis}
The goal of deformation analysis using the X-ray images is to quantify the  deformation of the flexible region of FPS to better understand its mechanics and effects during the pullout process. 
Figure \ref{fig:Xray} illustrates the see-through view of the sawbone samples during the experiments together with the results for the load-displacement and deformation analysis for an individual F9-3-0.5 and F9-4-0.5 screws. 
As can be observed in Fig. \ref{fig:Load-Disp} and Fig. \ref{fig:Xray}, the pitch of the FPS  plays a major role in the deformation behavior of the screw during the pullout test. An FPS with larger pitch (i.e., F9-4-0.5) demonstrate larger stiffness and therefore more pullout force and less oscillation and snap-back effect. Despite being more flexible due to a smaller pitch,  F9-3-0.5 FPS demonstrates longer time to fail compared with the other screw and a bigger PFZ. This can be correlated with the larger stored elastic energy of this screw compared with the other FPS.

\section{Conclusion and Future Work}
In order to mitigate the problems associated with pedicle screw pullout, we introduce a novel FPS capable of providing more than adequate pull out strength based on required standards \cite{ASTMF1839} while prolonging the displacement for failure and increased PFZ.  Overall, in this study and for the first time to our knowledge, we (i) proposed a design methodology for a unique semi-rigid, semi-flexible  FPS, (ii)  fabricated the FPS in titanium with two different pitches to validate the additive manufacturing process, (iii) utilizing a novel experimental setup synchronizing an X-ray machine with a tensile tester, validated the FPS pullout strength against current ASTM standards and discovered a novel resisting feature due to the elasticity of  the FPS, and (iv) analyzed the deformation behavior of the implanted part of the FPS hidden inside the Sawbone samples. In the future, we will further validate the fixation strength of  FPS by performing pullout tests on Sawbone samples and cadaveric specimens containing an FPS implanted in J-shape trajectories pre-drilled using our steerable drilling robotic system \cite{Sharma2023ACT,Sharma2023ANC,Sharma2023TowardsBD,Sharma_ismr2024,Sharma2024ABR}. We also plan to  make finite element models to optimally design and manufacture various versions of FPS.

\addtolength{\textheight}{-2cm}   



\bibliographystyle{IEEEtran}
\bibliography{root}

\end{document}